\documentclass[sn-mathphys-num]{sn-jnl}
\usepackage{graphicx}%
\usepackage{multirow}%
\usepackage{amsmath,amssymb,amsfonts}%
\usepackage{amsthm}%
\usepackage{mathrsfs}%
\usepackage[title]{appendix}%
\usepackage{xcolor}%
\usepackage{textcomp}%
\usepackage{manyfoot}%
\usepackage{booktabs}%
\usepackage{algorithm}%
\usepackage{algorithmicx}%
\usepackage{algpseudocode}%
\usepackage{listings}%
\usepackage{bm}
\usepackage{gensymb}

\theoremstyle{thmstyleone}

\theoremstyle{thmstyletwo}

\theoremstyle{thmstylethree}

\begin{document}

\title[Article Title]{Increasing the Task Flexibility of Heavy-Duty Manipulators Using Visual 6D Pose Estimation of Objects}
\author*[1]{\fnm{Petri} \sur{M{\"a}kinen}}\email{petri.makinen@tuni.fi}
\author[1]{\fnm{Pauli} \sur{Mustalahti}}\email{pauli.mustalahti@tuni.fi}
\author[2]{\fnm{Tuomo} \sur{Kivel{\"a}}}\email{tuomo.kivela@sandvik.com}
\author[1]{\fnm{Jouni} \sur{Mattila}}\email{jouni.mattila@tuni.fi}

\affil*[1]{\orgdiv{Automation Technology and Mechanical Engineering}, \orgname{Tampere University}, \orgaddress{\city{Tampere}, \country{Finland}}}
\affil[2]{\orgname{Sandvik Mining and Construction Oy}, \orgaddress{\city{Tampere}, \country{Finland}}}

\abstract{Recent advances in visual 6D pose estimation of objects using deep neural networks have enabled novel ways of vision-based control for heavy-duty robotic applications. In this study, we present a pipeline for the precise tool positioning of heavy-duty, long-reach (HDLR) manipulators using advanced machine vision. A camera is utilized in the so-called eye-in-hand configuration to estimate directly the poses of a tool and a target object of interest (OOI). Based on the pose error between the tool and the target, along with motion-based calibration between the camera and the robot, precise tool positioning can be reliably achieved using conventional robotic modeling and control methods prevalent in the industry. The proposed methodology comprises orientation and position alignment based on the visually estimated OOI poses, whereas camera-to-robot calibration is conducted based on motion utilizing visual SLAM. The methods seek to avert the inaccuracies resulting from rigid-body--based kinematics of structurally flexible HDLR manipulators via image-based algorithms. To train deep neural networks for OOI pose estimation, only synthetic data are utilized. The methods are validated in a real-world setting using an HDLR manipulator with a 5 m reach. The experimental results demonstrate that an image-based average tool positioning error of less than 2 mm along the non-depth axes is achieved, which facilitates a new way to increase the task flexibility and automation level of non-rigid HDLR manipulators.}


\maketitle

\section{Introduction}
The heavy-duty mobile machinery industry is evolving toward increased automation levels and autonomous operations, which offer potential benefits, such as increased safety and productivity \cite{lopes2018benefits}. This is empowered by the ongoing digital transformation and the adoption of so-called 4.0 technologies, which include automation and robotics \cite{sanchez2020innovation}. Heavy-duty machines are used in various sectors, including mining, construction, forestry, and material handling. Many of these machines are equipped with one or more heavy-duty, long-reach (HDLR) manipulators. While the mathematical foundations for modeling and the control methods of HDLR manipulators are mostly equivalent to those employed in conventional industrial robots found on factory floors, the unique characteristics of HDLR manipulators present significant challenges in terms of precise tool control. The primary challenge arises from the fact that HDLR manipulators are subject to considerable structural flexibility, especially bending in the direction of gravity. Unlike conventional industrial robots produced in bulk, HDLR manipulators are not rigid-bodied in practice. Consequently, using rigid-body--based modeling and control methods leads to compromised accuracy in tool center point (TCP) positioning. For fully automated operations, however, many work tasks require precise TCP positioning, which is a problem yet to be completely solved for HDLR manipulators in challenging environments.

In recent years, advances in imaging algorithms and deep neural networks have propelled scientific research forward. Notably, the field of visual 6D pose estimation of objects has gained prominence. This estimation can be based on either RGB or RGB-D imaging, with the latter consistently outperforming in BOP: benchmark for 6D object pose estimation \cite{hodan2018bop} leaderboards. Researchers often classify related methods into two categories: instance level and category level \cite{hoque2021comprehensive,costanzo2023non}. Instance-level methods rely on accurate 3D CAD models and extensive data covering various object poses in images. An object detector \cite{zaidi2022survey} is typically used first to find the region of interest (bounding box) of the object before the pose estimation. By contrast, category-level methods aim to generalize across unseen objects without relying on specific object models. Presently, instance-level methods yield the most accurate pose estimates and are best suited toward practical industrial applications \cite{guan2024survey}. Moreover, deep neural networks utilizing RGB-D data (combining color images and depth maps) outperform those trained solely on RGB or point cloud data, although many proposed methods utilize the depth map only for pose refinement, which is an additional step commenced after an initial RGB-based pose estimation. As uncovered in \cite{bauer2024challenges}, most research aims to maximize the performance on BOP benchmark datasets, whereas practical robotic systems and their requirements are not often directly considered. Furthermore, acquiring real-world data with ground-truth poses is very challenging because of annotation complexities. Thus, synthetic data are essential for network training, especially in practical applications.

The development of modern computer vision applications in general has created a need for annotated training data beyond large-scale public datasets. To address this problem, synthetic image data have become essential in acquiring data in a fast and cheap way \cite{man2022review}. Software related to 3D modeling and game engines can be used to implement a virtual environment. Although a reality gap \cite{tobin2017domain} exists when a network trained on synthetic data is deployed to the real world, the careful selection of the employed methodologies can help minimize this domain gap. Studies have shown that RGB-D trained networks generalize better to the real-world than RGB trained networks do, that data augmentation is essential when training with synthetic data, and that photo-realism and physically based rendering help in bridging the gap between synthetic training data and the real world \cite{hodavn2019photorealistic,pitteri2019object,tremblay2018deep,wang2019densefusion}. While the accuracy of pose estimation networks based on known objects has started to saturate, the refresh rate is one aspect that requires improvement \cite{hodan2023bop}. The current trend is also toward large-scale, pre-trained foundation models \cite{liu2024few,firoozi2023foundation}, which are intended to generalize for any unseen object. One option to overcome the issue of a low refresh rate is to switch to pose tracking after an initial pose estimate is obtained. The recent FoundationPose \cite{wen2024foundationpose} presented this methodology and reported a 32 Hz refresh rate for the pose tracking of one object. However, despite the higher refresh rate during the pose tracking thread, the overall performance and reliability in practical applications remain ambiguous. If the pose tracking deteriorates, a new pose initialization that is computationally costly, especially for a generalized network, is required.

Sensing and visual perception are both essential components of autonomous machines \cite{machado2021autonomous}, as machine vision enables visual recognition and decision making, leading to increased task flexibility and automation level. In general, emerging technologies related to intelligence, such as machine learning and machine vision, have attracted much attention in the field of heavy-duty machines. However, ongoing research is still mostly in a proof-of-concept phase \cite{khan2022overview}. To assess technological advancement, the technology readiness level (TRL) has been presented \cite{mankins1995technology}. Although originally proposed for flight and space applications, TRL can also be applied to other fields. It attempts to systematically depict the maturity of a particular technology. A \textit{proof of concept} translates to TRL 3, which is on the low end of the scale, further demonstrating the early-stage development of the heavy-duty machine sector with emerging technologies.

Vision-based control, or visual servoing, in the classical context of robotics is categorized into pose-based visual servoing (PBVS) \cite{wilson1996relative}, image-based visual servoing (IBVS) \cite{chaumette2007potential}, and hybrid systems \cite{hafez2008hybrid} that mix PBVS and IBVS. Pose-based visual servoing utilizes pose information in 3D space and is based on estimating the static transformation matrix from the camera frame to the robot's frame. Image-based visual servoing is based on computing an image Jacobian that maps pixel velocities to the camera's motion. Thus, IBVS is mostly utilized for control in a 2D plane of motion, although some research has been conducted on IBVS that attempts to utilize 3D features. Moreover, visual servo systems are mostly categorized into two configuration types: eye in hand and eye to hand \cite{hutchinson1996tutorial}. In the former, the camera is mounted at the end of the manipulator, while in the latter, the camera is fixed in the workspace (or on the mobile platform the manipulator is attached to). The eye-in-hand method provides more precise measurements, while the eye-to-hand method has a wider view of the environment \cite{makinen2020redundancy}. For HDLR manipulators, pose basedness is perceived as the best approach for vision-based control, but non-rigid structures make computing an accurate extrinsic calibration very challenging. Furthermore, extrinsic calibration using a checkerboard is not realistic for HDLR manipulators in challenging environments. Thus, to achieve precise tool positioning accuracy, this work utilizes motion-based calibration while computing vision-based pose errors for control in 3D Cartesian space. To achieve high precision with a camera, the sensor should be placed near the objects of interest (OOIs), making the eye-in-hand configuration most suitable.

In this study, the objective was the precise positioning of the TCP of an HDLR manipulator to a target OOI using advanced machine vision. In the context of HDLR manipulators with higher tolerances than industrial robots, a minimum positioning accuracy of $\pm$5 mm is desired. It is assumed that 3D CAD models of the OOI are available. To achieve precise tool positioning, motion-based calibration is performed to find the extrinsic relation between the camera and the HDLR manipulator. The method takes advantage of VO/SLAM (visual odometry/simultaneous localization and mapping) to estimate the camera pose trajectory. In our previous study \cite{makinen2024vision}, we presented a similar approach using fiducial markers as proof of concept. However, using external markers is not a realistic approach for HDLR manipulators in unknown environments. Thus, this study focuses on extending the methods of our prior research into practical relevance by utilizing deep neural networks for visual pose estimation of OOIs. The contributions of the present study are as follows: i) A complete pipeline for precise TCP positioning for non-rigid HDLR manipulators in OOI-focused applications using deep neural networks is presented and discussed; ii) the coarse alignment during the camera-to-robot calibration step is conducted in a global manner; and iii) a minimal path is introduced for motion-based calibration. The proposed methodology is validated in a laboratory setting, relating to TRL 4, using an HDLR manipulator with a 5 m reach and a camera in the eye-in-hand configuration. Notably, pose estimation networks for real-world OOIs are trained using synthetic data only. The experimental results demonstrate that an excellent TCP positioning accuracy is reliably achieved, which shows significant potential in enabling increased flexibility for automated tasks with HDLR manipulators, as advanced machine vision can be utilized to vary the control targets in a flexible manner.

The rest of the paper is organized as follows. Section 2 describes related preliminaries, Section 3 presents the methods, Section 4 discusses the implementation details, Section 5 presents the experimental results, and, finally, Section 6 concludes the paper.

\section{Modeling and Control of a Serial-Link Manipulator}
\label{sec:2}
The pose $\mathbf{x} \in \mathbb{R}^6$ represents the position and orientation of a robotic manipulator's TCP relative to its base frame within the operational space. The mapping from the operational space to the joint space is defined by forward kinematics, which requires knowledge of the joint variables $\mathbf{q} \in \mathbb{R}^n$. The forward kinematic equation is written as
\begin{equation}
	\label{eq:kin0}
	\mathbf{x} = 
	\begin{bmatrix}
		\mathbf{p} \\
		\bm{\theta}
	\end{bmatrix} =
	\mathbf{f}(\mathbf{q}), \quad
	\mathbf{q} = 
	\begin{bmatrix}
		q_1 \\
		\vdots \\
		q_n
	\end{bmatrix},
\end{equation}
where position $\mathbf{p} \in \mathbb{R}^3$ and orientation $\bm{\theta} \in \mathbb{R}^3$ define the TCP's pose. The TCP frame's rotation is given by Euler angles for minimal representation. The TCP's velocity relationship is:
\begin{equation}
	\label{eq:kin1}
	\dot{\mathbf{x}} =
	\begin{bmatrix}
		\dot{\mathbf{p}} \\
		\dot{\bm{\theta}}
	\end{bmatrix} = 
	\mathbf{J}(\mathbf{q})\dot{\mathbf{q}}.
\end{equation}
Here, $\mathbf{J}(\mathbf{q}) \in \mathbb{R}^{6\times n}$ denotes a Jacobian matrix that maps the joint velocities $\dot{\mathbf{q}} \in \mathbb{R}^n$ to the respective task space velocities $\dot{\mathbf{x}} \in \mathbb{R}^6$. To find joint velocities from known TCP velocities, the inverse Jacobian is applied:
\begin{equation}
	\label{eq:kin2}
	\dot{\mathbf{q}} = \mathbf{J}^{-1}(\mathbf{q})\dot{\mathbf{x}}.
\end{equation}
With knowledge of the desired TCP position $\mathbf{p}_d$ and orientation $\bm{\theta}_d$, along with the respective desired velocities $\dot{\mathbf{p}}_d$ and $\dot{\bm{\theta}}_d$, the desired joint velocities $\dot{\mathbf{q}}_d$ are obtained by modifying Eq.~\eqref{eq:kin2}:
\begin{equation}
	\label{eq:kin3}
	\dot{\mathbf{q}}_d = \mathbf{J}^{-1}(\mathbf{q})
	\begin{bmatrix}
		\dot{\mathbf{p}}_d + \mathbf{K}_p ({\mathbf{p}}_d - {\mathbf{p}}) \\
		\dot{\bm{\theta}}_d + \mathbf{K}_{\theta} \delta \mathbf{r}
	\end{bmatrix}.
\end{equation}
Here, $\mathbf{K}_p$ and $\mathbf{K}_{\theta}$ are the control gains for position and orientation feedback, respectively. The orientation error $\delta \mathbf{r}$ is represented by quaternions. The desired joint positions $\mathbf{q}_d$ are then obtained by integrating
\begin{equation}
	\label{eq:kin4}
	\mathbf{q}_d = \int \mathbf{J}^{-1}(\mathbf{q}) \dot{\mathbf{x}}_d dt.
\end{equation}
Finally, the control input vector $\mathbf{u}$ is defined as
\begin{equation}
	\label{eq:kin5}
	\mathbf{u} = \mathbf{K}_v (\mathbf{q}_d - \mathbf{q}),
\end{equation}
with $\mathbf{K}_v$ containing the joint control gains.

The TCP pose in Eq.~\eqref{eq:kin0} can also be expressed as a transformation matrix, comprising a rotation matrix and a translation vector. The forward kinematic model of a serial-link manipulator can be formulated using the well-known Denavit--Hartenberg (DH) convention. The rigid transformation relating the base of the manipulator to its TCP using DH parameters is computed as
\begin{equation}
	\label{eq:Tenc}
	^{\mathbf{B}}\mathbf{T}_{\mathbf{T}} = \prod_{i=1}^j \mathbf{T}_{i},
\end{equation}
where $j$ is the number of joints, and $\mathbf{T}_{i}$ denotes the joint-specific transformation matrices, formulated as
\begin{equation}
	\label{eq:gentranf}
	\mathbf{T}_i = 
	\begin{bmatrix}
		c\theta_i & -s\theta_i c\alpha_i & s\theta_i s\alpha_i & a_i c\theta_i \\
		s\theta_i & c\theta_i c\alpha_i & -c\theta_i s\alpha_i & a_i s\theta_i \\
		0 & s\alpha_i & c\alpha_i & d_i \\
		0 & 0 & 0 & 1
	\end{bmatrix},
\end{equation}
while using the DH parameters ($\theta_i, d_i, \alpha_i, a_i$) of the $i^{th}$ joint. Moreover, $\sin$ is abbreviated with $s$ and $\cos$ with $c$.

\section{Methods}
Considering the state of the art in visual 6D pose estimation of objects and its practical applicability for industrial purposes, an instance-level--based method is adopted. Such methods typically require a preceding object detector network, as the pose estimation network utilizes the cropped region of interest containing the OOI as an input. This section details the methods used for synthetic dataset generation, visual object detection, visual 6D pose estimation, and vision-based control of HDLR manipulators.

\begin{figure}[htbp]
	\centerline{\includegraphics[width=0.90\textwidth]{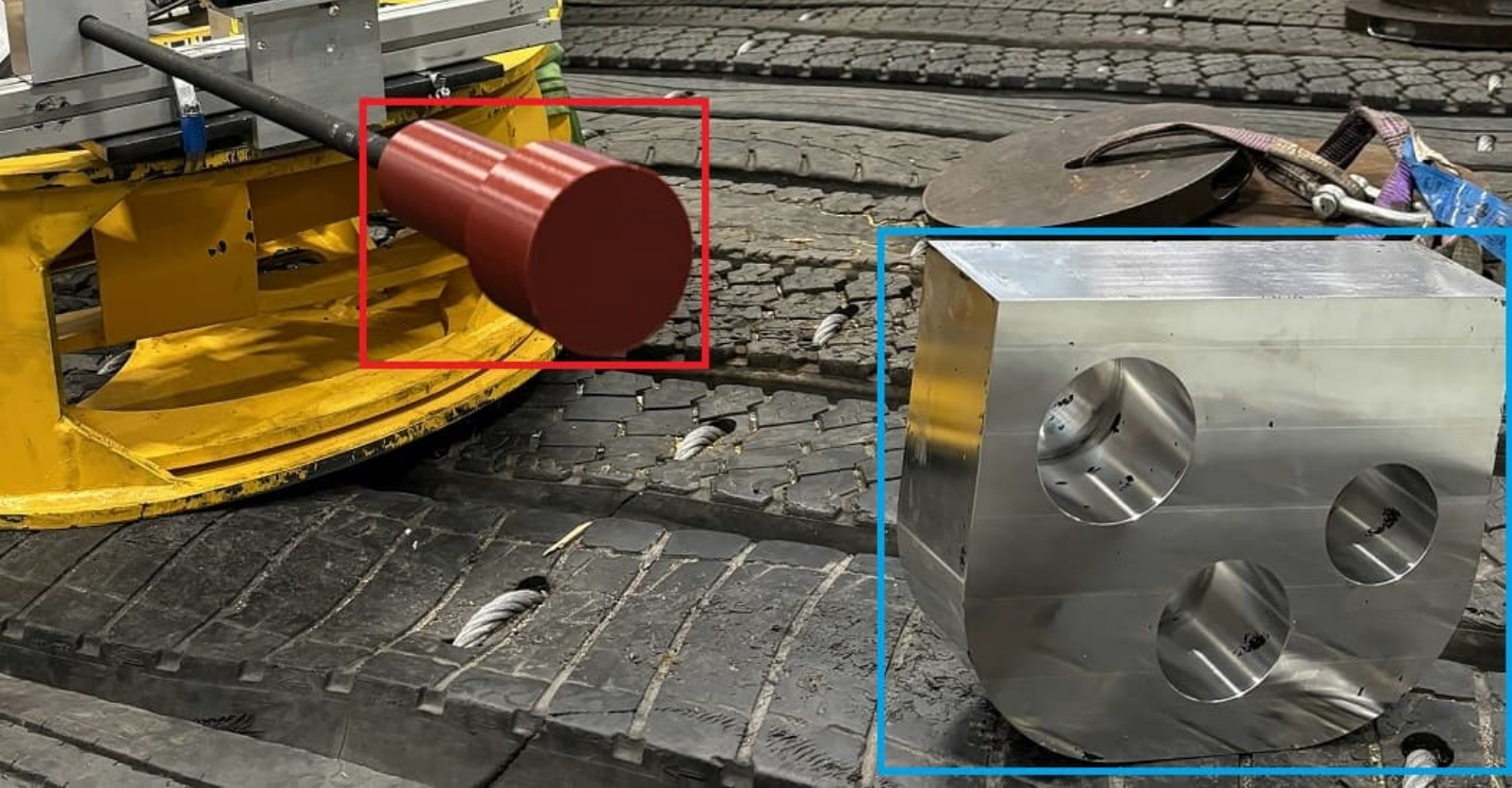}}
	\caption{The two OOIs used in this work. The mock-up tool attached to the manipulator is highlighted with the red bounding box, whereas the larger target OOI is highlighted with the blue bounding box.}
	\label{fig:objects}
\end{figure}

Two mock-up objects, shown in Fig.~\ref{fig:objects}, were considered in this work. The first OOI is a 3D printed peg with continuous symmetry. The second OOI is a slab made of aluminum with holes of varying sizes. The underlying use case is an industry-related insertion task, so the aim is to position the peg to one of the holes in the slab. The hole positions were mapped with respect to the object's base frame using the known geometry.

\subsection{Synthetic Data Generation}
To train an object detector network and pose estimation networks, a dataset is first required. To generate a synthetic dataset with domain randomization and with annotations according to the standard BOP format, the BOP toolkit and BlenderProc4BOP \cite{Denninger2023} were utilized. The custom dataset included the two OOIs and comprised approximately 50k images with a 1280$\times$720 resolution. The OOI and camera poses were randomized, but they were constrained to be realistic in compliance with the application. It was assumed that both OOIs are roughly facing the camera and that they are upright. Thus, the total amount of OOI poses that needed to be covered in the dataset was significantly reduced because of the real-world application considered. The larger OOI also has poses with discrete symmetries that were not considered because of the assumptions. These application-specific constraints are later used to perform consistency checks on the estimated OOI poses before forwarding them to the control system.

\begin{figure}[htbp]
	\centerline{\includegraphics[width=0.90\textwidth]{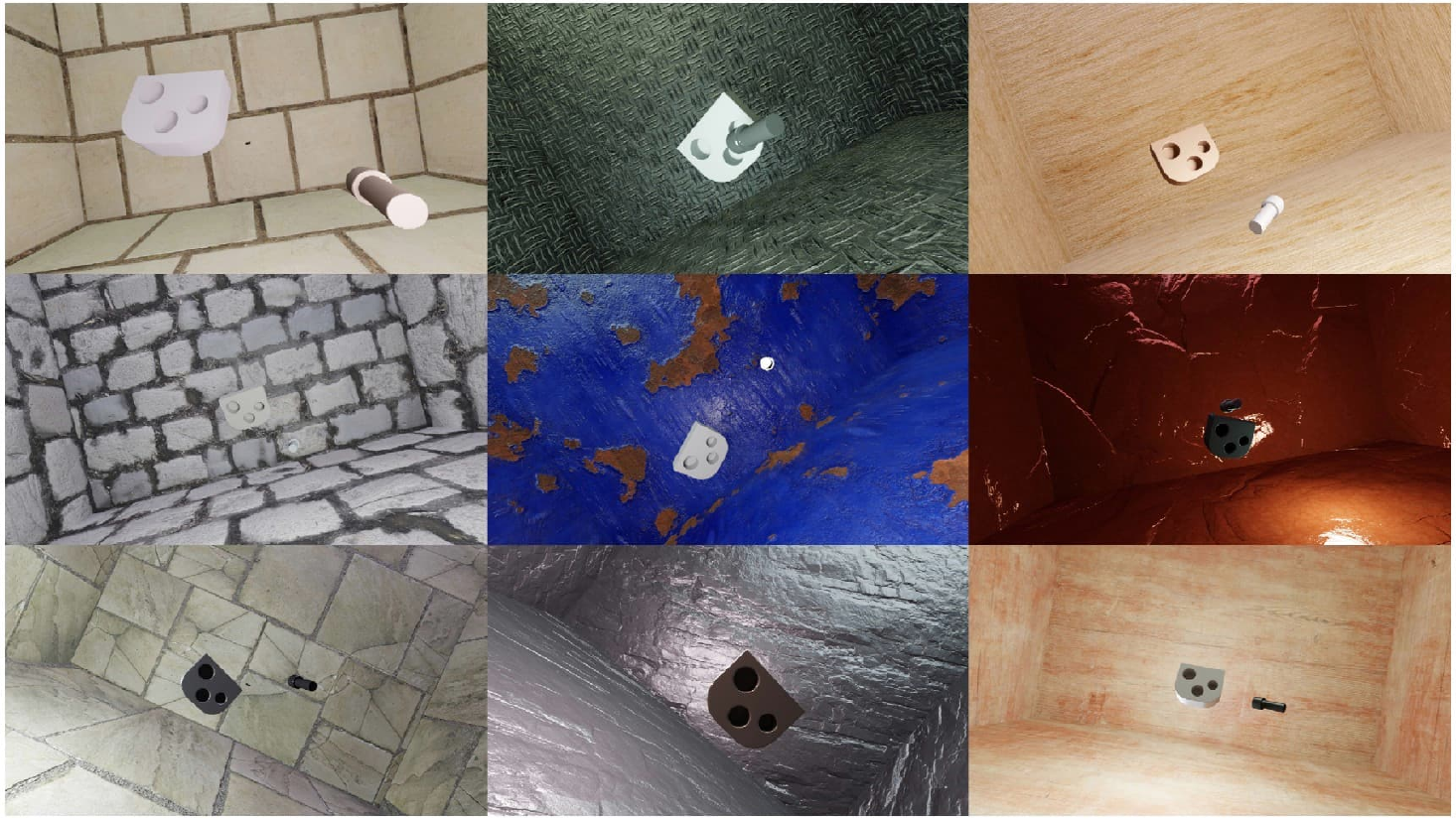}}
	\caption{Example color images of the synthetic dataset generated with BlenderProc4BOP.}
	\label{fig:pbrdata}
\end{figure}

The scene was a room with randomized background and lighting, with the OOI surfaces also randomized with a set of metallic textures. The distance between the two OOIs and the distance between an OOI and the camera were constrained to be uniform with the target application. Images from the generated dataset are illustrated in Fig.~\ref{fig:pbrdata}.

\subsection{Object Detection}
A YOLOv7 \cite{wang2023yolov7} visual object detector was trained for the two OOIs by utilizing the pre-trained weights provided by the authors. The network was fine-tuned with the custom synthetic dataset for 20 epochs. While the resulting object detector performed well for the synthetic data, real-world detections were not up to the standard. Therefore, an additional real-world dataset was constructed.

\begin{figure}[htbp]
	\centerline{\includegraphics[width=0.90\textwidth]{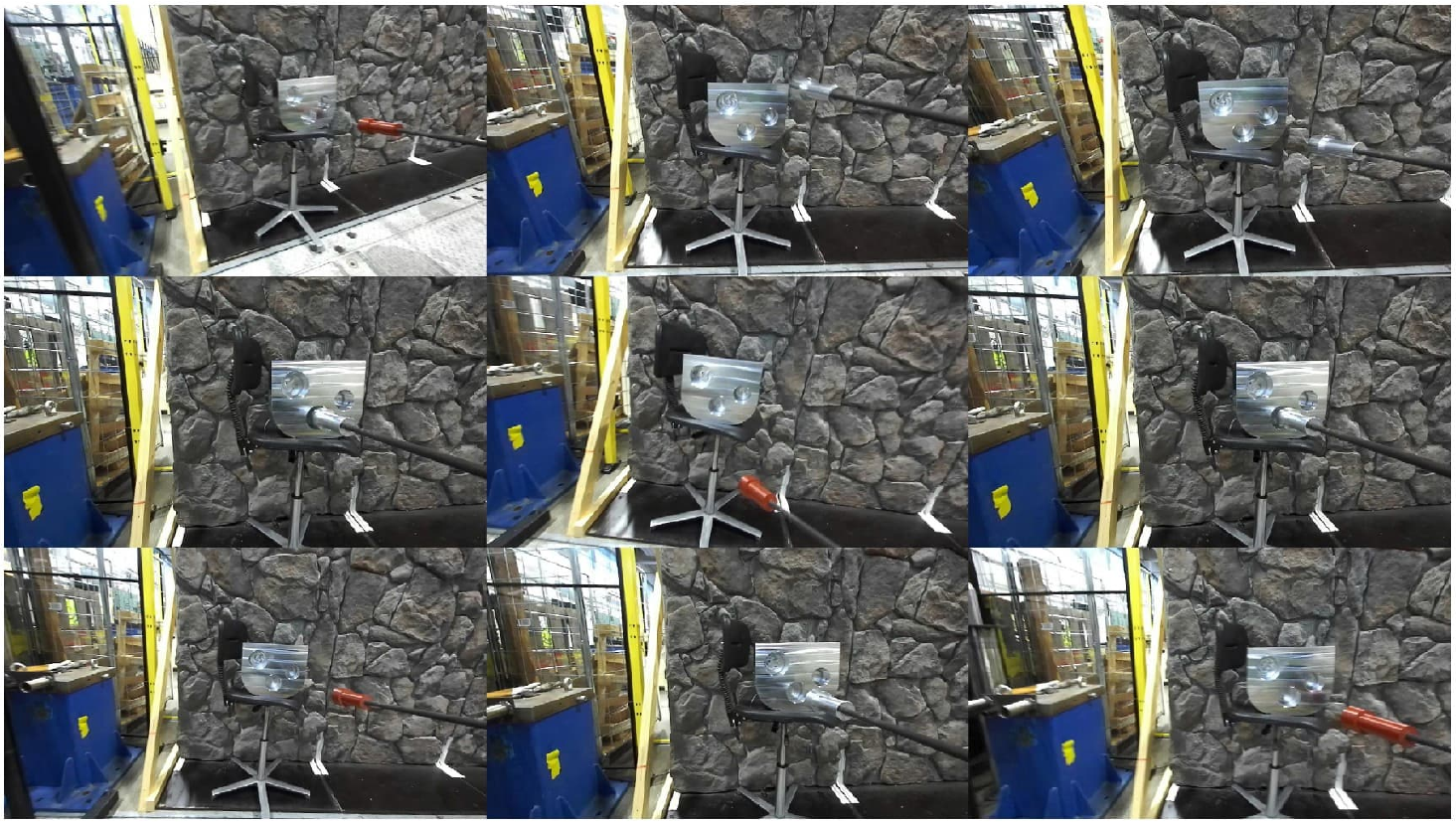}}
	\caption{Examples of the real-world dataset used for fine-tuning the visual object detector.}
	\label{fig:realdata}
\end{figure}

Videos comprising approximately 3k frames were semi-automatically labeled with rectangular bounding boxes for each OOI using MATLAB's Video Labeler tool. The object detector was then fine-tuned further with the real-world dataset for 14 epochs, which resulted in good real-world performance. Fig.~\ref{fig:realdata} illustrates images from the real-world dataset. Roboflow \cite{roboflow} was utilized in modifying the generated datasets for YOLOv7 compliance.

\subsection{Pose Estimation}
For visual 6D pose estimation of the two OOIs, ZebraPose \cite{su2022zebrapose} was employed. Based on BOP benchmark leaderboards, it has high accuracy with relatively low computation time. It is an instance-level method using coarse-to-fine surface encoding for 6D pose estimation of an OOI from an RGB image. The instance levelness implies that the pose is learned by showing the network each individual pose separately. Consequently, such networks do not generalize to unseen poses or unseen objects, but for a known OOI, high accuracy is achievable. The downsides are that a network is trained for each OOI separately and that the refresh rate is not real time. Many methods tend to utilize only the RGB image when learning the pose, while the depth map is utilized for the optional step of pose refinement, which is often computationally costly. Additional pose refinement was not used in this work.

A ZebraPose network was trained for each OOI using the generated synthetic dataset. The EfficientNetB4 \cite{tan2019efficientnet} backbone was used with Adam optimizer and a learning rate of $1e-4$. During training, the OOI poses were solved using Progressive-X \cite{barath2019progressive}, whereas during inference, the poses were solved using RANSAC/PnP (random sampling consensus/perspective-n-point), which was more than two times faster compared to the Progressive-X method. Both networks were trained for approximately 800k iterations using a batch size of 16. Training a single network took approximately a week on an NVIDIA RTX 6000. A detector file including the OOI bounding boxes was generated for the custom test data using the fine-tuned YOLOv7 object detector.

\subsection{Vision-based Control}

\subsubsection{Alignment Using OOI Orientations}
\label{subsubsec:oa}
The frames of the two OOIs are first aligned so that the tool OOI is facing the target OOI. The coordinate frames were defined in the 3D CAD modeling phase so that they are aligned in the desired configuration. Let the visually estimated pose of the OOI attached to the manipulator be denoted by a transformation matrix comprising a rotation matrix and a translation vector, $^{\mathbf{C}}\mathbf{T}_{\mathbf{O1}}$. Let the visually estimated pose of the target OOI be denoted by $^{\mathbf{C}}\mathbf{T}_{\mathbf{O2}}$, where $\mathbf{C}$ represents the camera frame. The rotation between the two OOI frames is then computed as a quaternion difference,
\begin{equation}
	\label{eq:rotdiff}
	\mathbf{q}_{\Delta} = \mathbf{q}_{O2} \mathbf{q}^{-1}_{O1},
\end{equation}
where $\mathbf{q}_{O2}$ and $\mathbf{q}_{O1}$ denote the OOI orientations expressed in quaternion. Then, a new reference pose $\mathbf{x}_{ref}$ for the manipulator is computed, expressed as a transformation matrix as
\begin{equation}
	\label{eq:rotdiff2} 
	\mathbf{T}_{\text{ref}} = {^{\mathbf{B}}\mathbf{T}_{\mathbf{T}}} {^{\mathbf{C}}\mathbf{T}_{\Delta}}
	={^{\mathbf{B}}\mathbf{T}_{\mathbf{T}}}
	\begin{bmatrix}
		{^{\mathbf{C}}\mathbf{R}_{\Delta}} & \mathbf{0}^{\text{T}} \\
		\mathbf{0} & 1
	\end{bmatrix},
\end{equation}
where ${^{\mathbf{C}}\mathbf{R}_{\Delta}} \in \mathbb{R}^{3\times3}$ represents the rotation obtained using quaternion difference, $\mathbf{0} \in \mathbb{R}^{1\times3}$, and ${^{\mathbf{B}}\mathbf{T}_{\mathbf{T}}}$ is obtained using Eq.~\eqref{eq:Tenc}.

\subsubsection{Motion-based Calibration}
\label{subsubsec:calib}
The proposed methodology is related to the PBVS in 3D Cartesian space. Thus, the description of the transformation from the camera frame to the TCP frame is required to enable vision-based control. For non-rigid HDLR manipulators, this extrinsic relation is obtained using motion-based calibration, in which the transformation is computed \textit{locally} by maintaining the TCP orientation and then utilizing trajectory matching between the camera pose and the joint encoder-based TCP pose (formulated using forward kinematics). The transformation between a manipulator's base frame and its TCP frame can be expressed using unit orientation vectors ($\mathbf{n}, \ \mathbf{s}, \ \mathbf{a}$) and the position vector $\mathbf{p}$ \cite{sciavicco2001modelling}:
\begin{equation}
	\label{eq:clc0}
	^{\mathbf{B}}\mathbf{T}_{\mathbf{T}}(\mathbf{q}) = 
	\begin{bmatrix}
		^{\mathbf{B}}\mathbf{n}_{\mathbf{T}}(\mathbf{q}) & ^{\mathbf{B}}\mathbf{s}_{\mathbf{T}}(\mathbf{q}) & ^{\mathbf{B}}\mathbf{a}_{\mathbf{T}}(\mathbf{q}) & ^{\mathbf{B}}\mathbf{p}_{\mathbf{T}}(\mathbf{q}) \\
		0 & 0 & 0 & 1
	\end{bmatrix}.
\end{equation}
Next, a point-to-point path for motion-based calibration is formulated with the following equations:
\begin{align}
	\label{eq:psm1}
	x_{i+1} &= x_i + D_i \cos(\gamma_i) n_1 + D_i \sin(\gamma_i) s_1, \\
	\label{eq:psm2}
	y_{i+1} &= y_i + D_i \cos(\gamma_i) n_2 + D_i \sin(\gamma_i) s_2, \\
	\label{eq:psm3}
	z_{i+1} &= z_i + D_i \cos(\gamma_i) n_3 + D_i \sin(\gamma_i) s_3,
\end{align}
where $[x_1, y_1, z_1]^{\text{T}} = {^{\mathbf{B}}\mathbf{p}_{\mathbf{T}}}$, and the unit orientation vectors $\mathbf{n} \in \mathbb{R}^3$ and $\mathbf{s} \in \mathbb{R}^3$ are obtained from Eq.~\eqref{eq:clc0}. The path is designed as asymmetric to improve the trajectory matching outcome. One of the aims was to use a minimized path length. Therefore, the parameters used were $D_1 = 0.05$ m, $D_2 = 0.2$ m, $\gamma_1 = 0$, and $\gamma_2 = \pi/2$.

During the execution of the calibration path, the camera pose is estimated and recorded using a VO/SLAM algorithm, and the encoder-based TCP is also recorded. The rigid transform from the camera frame to the TCP frame is computed using point set (i.e., trajectory) matching, which comprises coarse frame alignment and subsequent fine matching using an iterative algorithm. The purpose of the coarse alignment is to align the two point sets roughly so that the following iterative algorithm has a very high likelihood of converging to a correct solution instead of a mirrored solution, for example.

Following the camera placement and kinematic planning, the coarse alignment was realized by first instantiating the recorded VO/SLAM points with the initial TCP orientation of the recorded data as follows:
\begin{equation}
	\label{eq:Rcfa}
	{^\text{rot}}\mathbf{T}_{\text{cfa}} = 
	\begin{bmatrix}
		\big({^{\mathbf{B}}\mathbf{R}_{\mathbf{T}}}\big)_{i=1} & \mathbf{0}^{\text{T}} \\
		\mathbf{0} & 1 \\
	\end{bmatrix}.
\end{equation}
Next, a coarse transformation for translation is formulated. For a point cloud $\mathbf{P}_C \in \mathbb{R}^{3\times N}$, the mass center is computed with
\begin{equation}
	\mathbf{P}_{C, center} = \frac{1}{N} \sum_{i=1}^{N} \mathbf{P}_{C, i},
\end{equation}
where $N$ is the length of the point set. For two point sets, the center difference is computed as
\begin{equation}
	\mathbf{P}_{\Delta} = \mathbf{P}_{C_2, center} - \mathbf{P}_{C_1, center}.
\end{equation}
The coarse transformation with respect to translation from the VO/SLAM point set to the encoder-based TCP point set is then formulated as
\begin{equation}
	\label{eq:Tcfa}
	{^\text{pos}}\mathbf{T}_{\text{cfa}} = 
	\begin{bmatrix}
		{\mathbf{I}^{3\times3}} & \mathbf{P}_{\Delta} \\
		\mathbf{0} & 1 \\
	\end{bmatrix},
\end{equation}
where ${\mathbf{I}}$ denotes an identity matrix. For iterative point set matching, a probabilistic method \cite{min2018robust} utilizing full 6 degrees-of-freedom (DOF) pose information was employed. The complete sequence of transformations to express a given VO/SLAM-based pose with respect to the manipulator's base frame is formulated as
\begin{equation}
	\label{eq:camTcalib}
	{^\mathbf{C}}\mathbf{T}_{\text{calib}} = \mathbf{T}_{\text{fm}}{^\text{pos}}\mathbf{T}_{\text{cfa}}
	{^\text{rot}}\mathbf{T}_{\text{cfa}} {^\mathbf{C}}\mathbf{T}_{\text{original}},
\end{equation}
where $\mathbf{T}_{\text{fm}}$ is the transformation resulting from iterative point set matching.

\subsubsection{Alignment Using OOI Positions}
\label{subsubsec:pos}
Using the motion-based calibration, comprising coarse rotation alignment $^{\text{rot}}\mathbf{T}_{\text{cfa}}$ and fine matching $\mathbf{T}_{\text{fm}}$, a position measurement in the camera frame is rotated to match the TCP frame as
\begin{equation}
	\label{eq:psm5}
	{^{\mathbf{C}}\mathbf{T}_{\Delta, calib}} = 
	\begin{bmatrix}
		\mathbf{R}_{\text{fm}} & \mathbf{0}^{\text{T}} \\
		\mathbf{0} & 1
	\end{bmatrix}
	\begin{bmatrix}
		\mathbf{R}_{\text{cfa}} & \mathbf{0}^{\text{T}} \\
		\mathbf{0} & 1
	\end{bmatrix}
	{^{\mathbf{C}}\mathbf{T}_{\Delta}},
\end{equation}
where ${^{\mathbf{C}}\mathbf{T}_{\Delta}}$ incorporates the positioning error between the two OOIs, measured in the camera frame. Note that because of the use of the camera-based error directly, this part only requires the rotation matrices for calibration. The positioning error is formulated as
\begin{equation}
	\label{eq:psm4}
	{^{\mathbf{C}}\mathbf{T}_{\Delta}} = 
	\begin{bmatrix}
		1 & 0 & 0 & {^{O2}}p_x-{^{O1}}p_x \\
		0 & 1 & 0 & {^{O2}}p_y-{^{O1}}p_y \\
		0 & 0 & 1 & {^{O2}}p_z-{^{O1}}p_z - z_o  \\
		0 & 0 & 0 & 1 
	\end{bmatrix},
\end{equation}
where $z_o$ is an offset between the depth parameters of the two OOIs. Considering Eqs.~\eqref{eq:psm5}--\eqref{eq:psm4} and maintaining the current TCP orientation, the new TCP pose reference that satisfies the vision-based control task is computed as
\begin{equation}
	\label{eq:psm6}
	\mathbf{T}_{\text{ref}} =
	\begin{bmatrix}
		{^{\mathbf{B}}\mathbf{R}_{\mathbf{T}}} & {^{\mathbf{C}}\mathbf{p}_{\mathbf{\Delta, calib}}} + {^{\mathbf{B}}\mathbf{p}_{\mathbf{T}}} \\
		\mathbf{0} & 1
	\end{bmatrix},
\end{equation}
from which the reference pose ${\mathbf{x}}_{ref} \in \mathbb{R}^6$ is extracted. The reference position can be updated using a new visual measurement as long as the TCP orientation is held, thus maintaining the validity of the motion-based calibration.

\section{Implementation Details}
\label{sec:4}
\begin{figure}[htbp]
	\centerline{\includegraphics[width=0.90\textwidth]{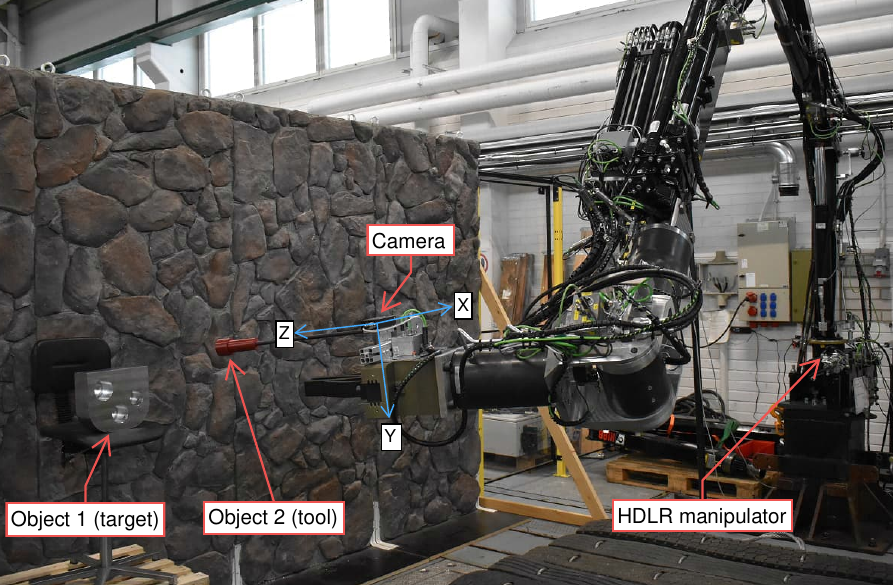}}
	\caption{The experimental setup comprising an HDLR manipulator with an eye-in-hand camera and two OOIs. The control objective was to position the tool OOI to one of the holes of the target OOI. The static mapping from each hole to the target's base frame was based on the known geometry.}
	\label{fig:expsetup}
\end{figure}

A laboratory-installed HDLR manipulator (HIAB033 articulated crane with an additional 3 DOF spherical wrist) with an approximately 5 m reach was used in the experiments, and it relates the results of this research to TRL 4. The experimental setup is illustrated in Fig.~\ref{fig:expsetup}, which also shows the two OOIs and the camera frame. The forward and inverse kinematic relations of the manipulator were formulated according to Section~\ref{sec:2}, with the final frame of the kinematic chain being unified with the camera frame.

Joint control was implemented using Eq.~\eqref{eq:kin5}, and the two most significant joints, lift and tilt, used PT-1 control:
\begin{equation}
	\label{eq:pt1}
	G(s) = \frac{K_p}{\tau s + 1},
\end{equation}
where $K_p$ is the gain, and $\tau$ denotes delay. This enables larger gain values compared to P-control, reducing static positioning errors when tracking performance is not a primary concern. The other joints, the rotation of the pillar and 3 DOF in the wrist, used P-controllers.

A rugged LIPSedge AE470 RGB-D camera, with known intrinsic parameters and in eye-in-hand configuration, was utilized for all visual measurements. To estimate the camera pose during the calibration path, ORB-SLAM3 \cite{campos2021orb} RGB-D was employed. The algorithm was ran at 640$\times$480@30FPS. The calibration path in Subsection~\ref{subsubsec:calib} was implemented in a point-to-point manner using quintic polynomials as in \cite{jazar10}.

For visual detection and 6D pose estimation of the two OOIs, the camera was ran at 1280$\times$720@30FPS. A reference TCP pose $\mathbf{x}_{ref}$ was first obtained using Eq.~\eqref{eq:rotdiff2}, or Eq.~\eqref{eq:psm6}, which was forwarded to the point-to-point trajectory generator. The inference time, including the object detector and two pose estimation networks with no visualization, was 0.306 s. This number was averaged over 292 samples. The minimum inference time was 0.265 s and the maximum time was 0.896 s. Thus, the vision-based control scheme was constricted to \textit{looking and then moving} (open-loop visual control) because of the non-real-time visual control updates. To reject any false positive OOI detections, only the ones with the highest confidence values for each OOI were forwarded to the respective pose estimation networks.

To increase the robustness and reliability of the vision-based control system, consistency checks were performed on the estimated OOI poses prior to forwarding them to the manipulator's control system. These checks were based on physical constraints and the expected OOI pose behavior, which are profoundly application specific. Furthermore, geometric moving average filtering \cite{gma} was applied to the estimated OOI poses to reduce noise. This type of filtering adds some delay to the signals. However, this is not a significant issue with an open-loop vision-based control scheme.

\begin{figure}[htbp]
	\centerline{\includegraphics[width=0.80\textwidth]{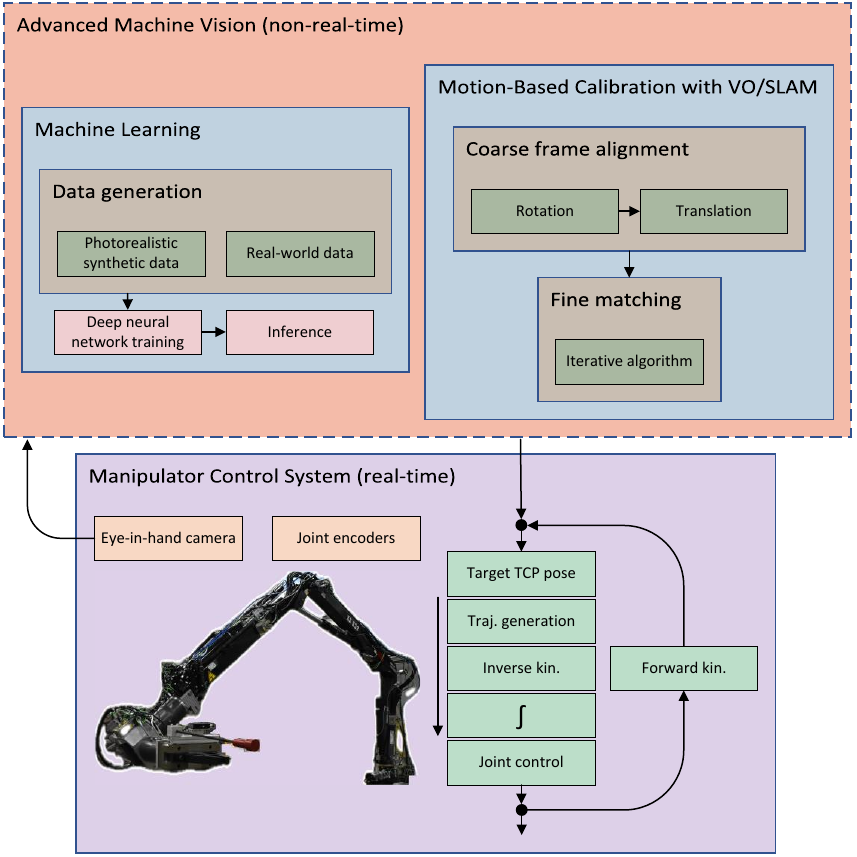}}
	\caption{The overall methodology for the precise TCP positioning of HDLR manipulators in OOI-focused applications. The advanced machine vision system comprises machine learning-related methods for visual OOI detection and pose estimation, along with VO/SLAM for motion-based calibration. The manipulator's real-time control system consists of the lower-level joint control guided by a given reference TCP pose.}
	\label{fig:chart}
\end{figure}

The vision-based algorithms, the VO/SLAM algorithm and visual OOI detection and pose estimation, were running on a separate Linux PC. All the pose data were transmitted to a Beckhoff real-time PC via UDP (user datagram protocol), on which the primary control system of the manipulator was running at a 1 ms sampling period. The overall methodology is illustrated in Fig.~\ref{fig:chart}, which combines advanced machine vision with industry-prevalent conventional robot modeling and control methods to achieve precise TCP positioning in OOI-focused tasks.

\section{Experimental Results}
To validate the presented methods, repeated experiments were conducted. The first step of each measurement was to perform the OOI alignment using estimated orientations, followed by the motion-based calibration, and, finally, the OOI positioning. The measurement results for each step are reported, and 10 measured cases are presented. The target OOI was moved in a different position in the workspace after each measurement, and the initial assumption was that the OOIs are in view of the camera.

Several evaluation metrics have been proposed to measure the results of visual 6D pose estimation of objects, such as visible surface discrepancy, maximum symmetric surface distance, and maximum symmetric projection distance \cite{hodavn2016evaluation,guan2024survey}. The BOP benchmark \cite{hodan2018bop} results measure the pose error by averaging over these three error functions. However, these metrics require ground-truth poses, which are not available for our real-world experiments. To investigate the correctness of the estimated OOI poses, visualization is a useful initial tool. The 3D models are rendered to the estimated pose into the respective image using the BOP toolkit. If the pose is estimated correctly, the rendered objects match the real-world objects' poses. To further examine the accuracy of the vision-based control system, the image-based position and orientation errors between the two OOIs are used. Ultimately, the underlying control objective was to precisely position the two OOIs with respect to each other. The specific error metrics and their results are detailed below.

\subsection{Alignment Using OOI Orientations}
\label{subsec:ooires}
\begin{figure}[htbp]
	\centerline{\includegraphics[width=0.90\textwidth]{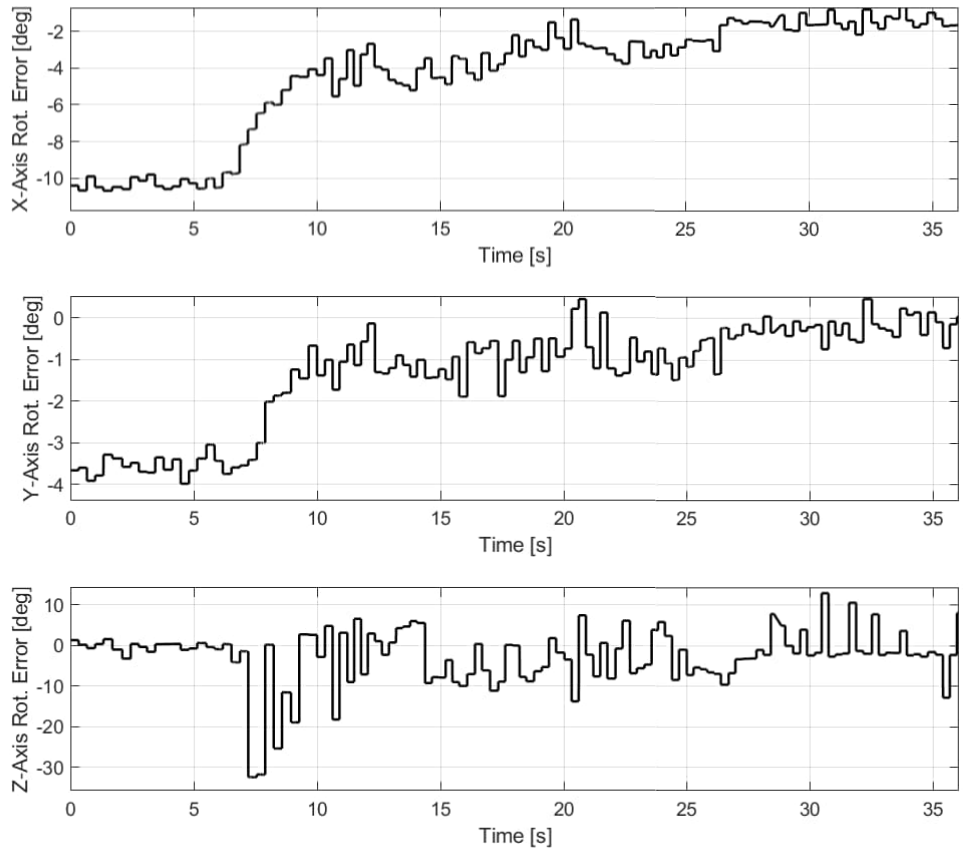}}
	\caption{An example result of the OOI alignment using orientations. The objective was to drive the orientation errors toward zeros in order to align the OOI frames. The absolute errors, expressed using Euler XYZ angles, were computed using the reference TCP orientation and the current TCP orientation.}
	\label{fig:exp_roterror}
\end{figure}

The first step in each measured case was to orient the tool OOI to face the target OOI in a perpendicular manner. Based on the CAD model designs, this implied simply aligning the OOI frames to match with one another. For the symmetrical tool, only two rotational axes are required. Thus, the last row (the symmetry axis) of the rotation matrices describing the orientation of each OOI were set to $[0, 0, 1]$. Employing the methods detailed in Subsection~\ref{subsubsec:oa}, the manipulator was then reoriented using the visual 6D pose estimates. With point-to-point trajectory generation, several consecutive reference poses computed from the visual feed were provided to the lower-level joint control system to reach sufficient accuracy with respect to the OOI orientations. Fig.~\ref{fig:exp_roterror} illustrates the orientation errors for one of the measured cases, with uniform results across all the measured cases. The reported errors were computed using the absolute error between the reference TCP orientation, extracted from Eq.~\eqref{eq:rotdiff2}, and the current TCP orientation, extracted from ${^{\mathbf{B}}\mathbf{T}_{\mathbf{T}}}$, expressed in Euler XYZ angles. The Z-angle error is also shown, although it is related to the symmetry axis that was unimportant in this case.

Table~\ref{Table_oa} lists the absolute orientation errors in each of the 10 measured cases. The reported numbers were averaged over the last 5 s of each dataset to reduce the effect of noise. The mean absolute errors for the two significant angles over all 10 measured cases were 1.09$^\circ$ for the X-angle and 0.52$^\circ$ for the Y-angle, which are sufficiently small errors for an HDLR manipulator. Across the 10 measured cases, there exists some variation between individual errors. The point-to-point looking and then moving vision-based control system was given several consecutive commands to \textit{converge} to a minimal orientation error. Specifically, three control updates for the TCP orientation were given in each measurement. For example, in Fig.~\ref{fig:exp_roterror}, the control commands were given at approximately 5 s, 14.5 s, and 24 s. The final errors were expected to include some variation, as the initial pose (i.e., the camera distance from the target OOI) of the TCP was not constant, and the target OOI was also moved around the workspace between the measured cases. Overall, the alignment using OOI orientations was conducted successfully and reliably in each case.
\begin{table}[h]
	\small\sf\centering
	\caption{Absolute orientation errors of the TCP, expressed in Euler XYZ angles and averaged over the last 5 s of each measurement. The Z-angle is associated with the unimportant symmetry axis.\label{Table_oa}}
	\begin{tabular}{llll}
		\toprule
		Case & X-angle [$^\circ$] & Y-angle [$^\circ$] & Z-angle [$^\circ$] \\
		\midrule
		1 & $0.89$ & $0.56$ & $6.39$ \\
		2 & $1.05$ & $0.63$ & $5.64$ \\
		3 & $1.16$ & $0.67$ & $9.28$ \\
		4 & $1.61$ & $0.34$ & $4.06$ \\
		5 & $0.43$ & $0.52$ & $13.03$ \\
		6 & $0.70$ & $0.25$ & $2.03$ \\
		7 & $1.04$ & $0.58$ & $5.89$ \\
		8 & $1.86$ & $0.44$ & $0.98$ \\
		9 & $0.68$ & $0.38$ & $4.88$ \\
		10 & $1.51$ & $0.87$ & $8.03$ \\
		\midrule
		Mean & $1.09$ & $0.52$ & $6.02$ \\
		\bottomrule
	\end{tabular}
\end{table}

\subsection{Motion-based Calibration with VO/SLAM}
\begin{figure}[htbp]
	\centerline{\includegraphics[width=0.60\textwidth]{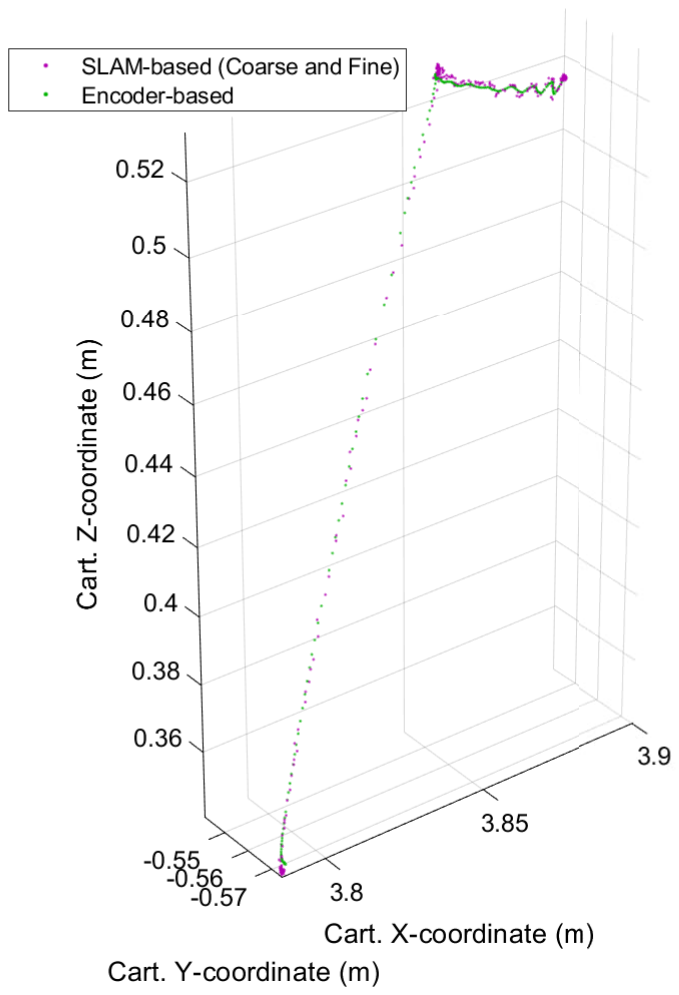}}
	\caption{The final result after iterative matching between the coarsely aligned SLAM points and the encoder-based TCP points.}
	\label{fig:exp_calib}
\end{figure}

Following the OOI alignment using the estimated orientations, the designed path for motion-based calibration was executed. The resulting VO/SLAM poses were first coarsely aligned to the encoder-based TCP pose trajectory using Eq.~\eqref{eq:Rcfa} and Eq.~\eqref{eq:Tcfa}. Then, the coarsely aligned SLAM poses and the encoder-based TCP poses were matched using the iterative method. The procedure of coarse and fine matching took approximately 0.4--0.5 s on an Intel i7-6700 CPU. The final trajectory matching result is computed using Eq.~\eqref{eq:camTcalib}. An example of this result is illustrated in Fig.~\ref{fig:exp_calib}, which shows that the SLAM points (after coarse and fine alignment) match near perfectly with the encoder-based TCP points, ensuring a successful calibration based on the motion.

Table~\ref{Table_calib} reports the mean and maximum absolute errors, resulting from the iterative matching, for the 10 measured cases. The matching result was uniform in each measured case, with a mean error less than 1 mm for the X-axis and Y-axis, and a mean error of slightly over 2 mm for the Z-axis. The respective maximum absolute errors were naturally larger but not significantly. Considering precise TCP positioning using OOI positions, the most important aspect of motion-based calibration is to obtain an accurate representation of the camera frame's orientation with respect to the manipulator's base frame, whereas the positional error resulting from the point set matching has no direct effect on the vision-based control task.

The success of the motion-based calibration pipeline, detailed in Subsection~\ref{subsubsec:calib}, is reliant on the performance of the VO/SLAM algorithm. In the laboratory setting (correlating with TRL 4), the utilized ORB-SLAM3 RGB-D was able to provide accurate pose trajectories in each direction with respect to the HDLR manipulator. However, for visual SLAM, it is required that the environment, although unknown and unstructured, has enough textured surfaces for feature extraction. If this is not the case, it may be possible to replace the visual SLAM with a light detection and ranging (LIDAR) SLAM algorithm. This has been shown to perform better in outdoor environments compared to visual SLAM \cite{ebadi2023present,zhao2024comprehensive}, albeit such a solution would necessitate an additional step of complexity in the form of LIDAR-to-camera calibration in this particular application.
\begin{table}[h]
	\small\sf\centering
	\caption{Mean and maximum absolute errors of point set matching between the coarsely aligned camera pose trajectory and the encoder-based TCP pose trajectory.\label{Table_calib}}
	\begin{tabular}{llll}
		\toprule
		Cases 1--10 & X-axis [mm] & Y-axis [mm] & Z-axis [mm] \\
		\midrule
		Mean & $0.99$ & $0.33$ & $2.05$ \\
		Max. & $1.30$ & $0.50$ & $3.80$ \\
		\bottomrule
	\end{tabular}
\end{table}

\subsection{Alignment Using OOI Positions}
\begin{figure}[htbp]
	\centerline{\includegraphics[width=0.90\textwidth]{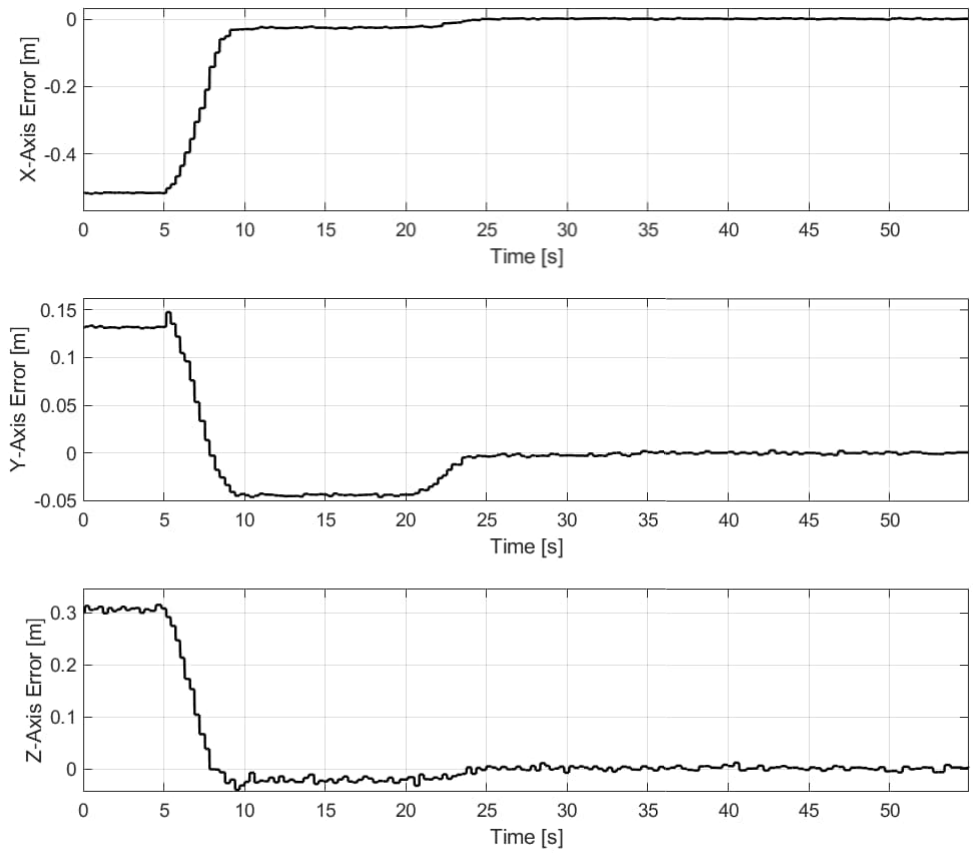}}
	\caption{An example result of OOI positioning. The objective was to drive the position errors toward zeros and position the tool OOI to a specific hole of the target OOI. The image-based absolute errors were computed using the estimated OOI poses.}
	\label{fig:exp_poserror}
\end{figure}

With the OOI orientations aligned, along with the motion-based calibration matrix available, the final step in the examined case is to position the OOI in the desired configuration using Eq.~\eqref{eq:psm6}. An example result of the OOI positioning is illustrated in Fig.~\ref{fig:exp_poserror}, which shows the image-based position errors between the two OOIs. Similarly with the orientations in Subsection~\ref{subsec:ooires}, several point-to-point commands were provided to the system to reduce the errors to a satisfactory range. In the example, the visual control commands were given at approximately 4 s, 19 s, and 31 s, which resulted in the desired tool positioning accuracy. It was also visually verified that the physical objects were indeed aligned.

Table~\ref{Table_poserr} reports the results for each of the 10 measured cases. As the aim was to drive the position errors to zeros, three to four visual control commands were provided to the control system depending on the case. The initial TCP position varied throughout the measurements, and some noise exists in the pose estimates. Consequently, the reported errors were averaged over the final 5 s of each data. The resulting errors across the measurements were uniform, with the mean position errors related to the X-axis and the Y-axis being less than 2 mm. The depth error along the Z-axis had a mean less than 6 mm. A larger error in the depth was expected, as it had the most uncertainty, especially considering RGB-based pose estimation. In applications related to HDLR manipulators with high forces, a minor depth-axis error is insignificant compared to the two other axes that determine the tool's alignment with reference to the target. However, collision with the target is not desired, so the tunable offset parameter was incorporated in Eq.~\eqref{eq:psm4}.

\begin{figure}[htbp]
	\centerline{\includegraphics[width=0.90\textwidth]{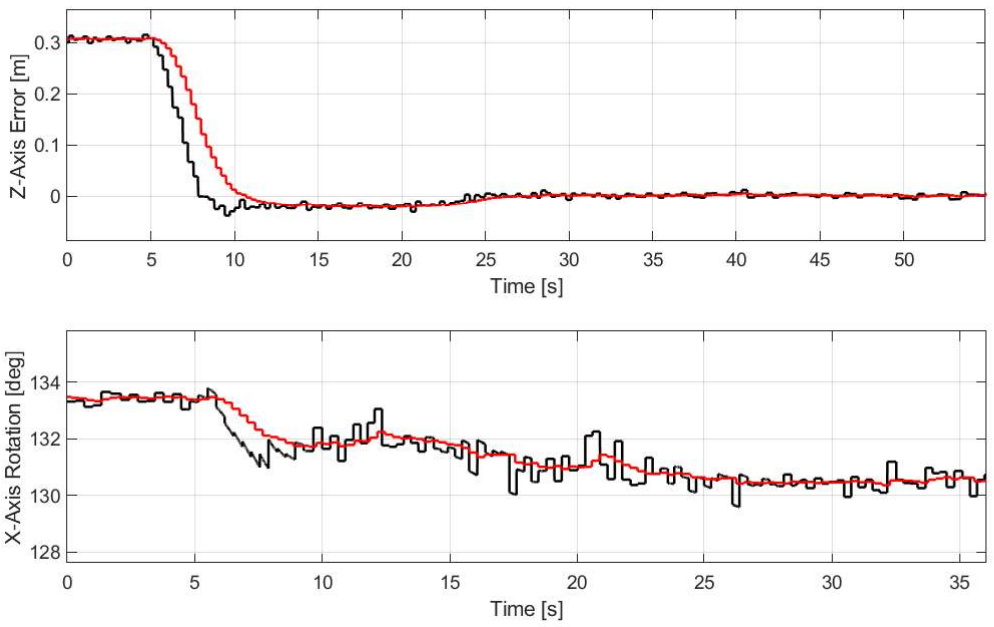}}
	\caption{An example of pose signals enhanced with geometric moving average filtering (red lines) compared to their respective raw signals (black lines). The upper figure shows the position error along the depth axis, and the bottom figure shows a rotation along the X-axis (in Euler XYZ angle format).}
	\label{fig:gma}
\end{figure}

As mentioned in Section~\ref{sec:4}, the OOI pose signal qualities were enhanced using geometric moving average filtering. Fig.~\ref{fig:gma} illustrates an enhanced position and orientation signal with their respective raw counterparts. The raw signals are direct pose outputs computed with RANSAC/PnP from the ZebraPose networks. While the raw poses were not of bad quality, the aim of filtering was to reduce any noise to achieve the most accurate positioning result. The filtering also reduced initial overshoot along the depth axis. Consequently, the filtering scheme induces signal delay, which, in this application, was acceptable.

\begin{figure}[htbp]
	\centerline{\includegraphics[width=0.90\textwidth]{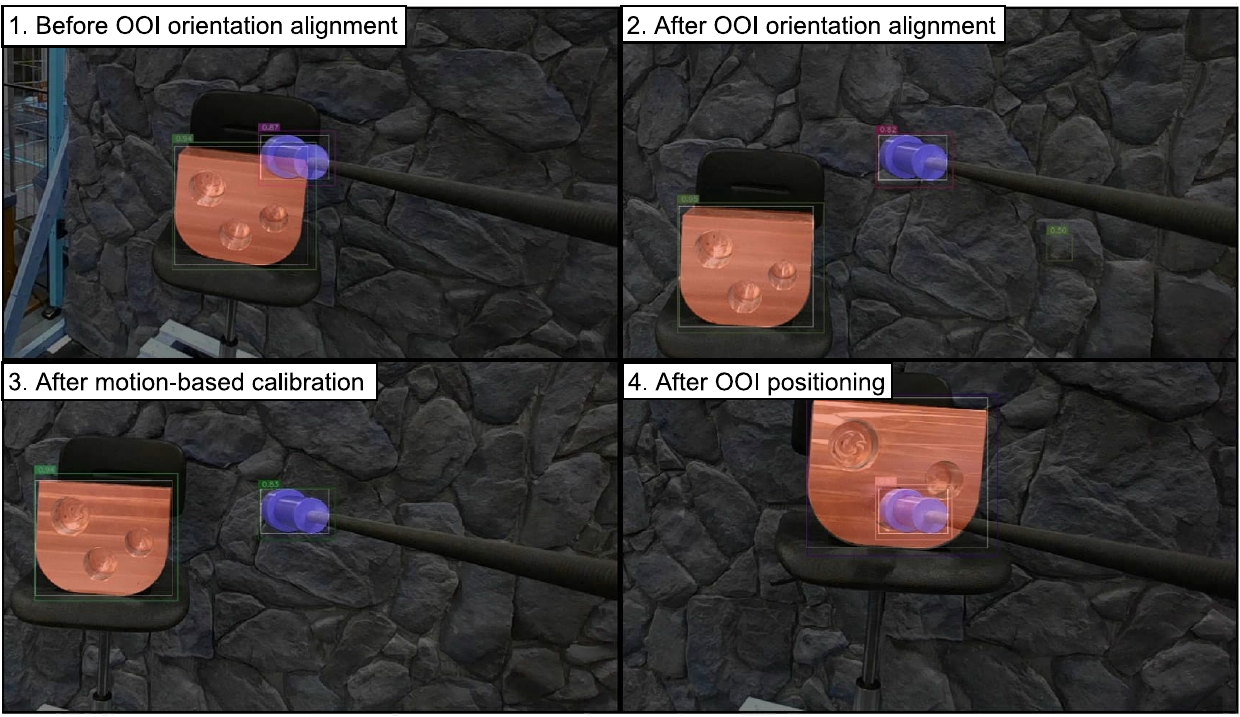}}
	\caption{Camera views before and after the OOI orientation alignment, after the calibration path, and after the final OOI positioning. The OOI models were rendered in the image to their estimated poses using the BOP toolkit. The original images have been cropped for illustration purposes.}
	\label{fig:expmontage}
\end{figure}

Finally, a visualization of the entire methodology is shown in Fig.~\ref{fig:expmontage}, which illustrates the camera point of view 1) at the initial pose, 2) after orientation alignment, 3) after the motion-based calibration, and 4) after final positioning to the middle hole, yielding an image-based positioning error of less than 2 mm with respect to the X-axis and Y-axis. The visualization was ran only separately, not during online experiments.
\begin{table}[h]
	\small\sf\centering
	\caption{Absolute position errors between the two OOIs after position alignment, measured in the camera frame and averaged over the last 5 s of each measurement.\label{Table_poserr}}
	\begin{tabular}{llll}
		\toprule
		Case & X-axis [mm] & Y-axis [mm] & Z-axis (depth) [mm] \\
		\midrule
		1 & $1.34$ & $1.40$ & $8.91$ \\
		2 & $2.03$ & $0.98$ & $4.39$ \\
		3 & $1.43$ & $1.90$ & $6.66$ \\
		4 & $1.16$ & $0.79$ & $3.21$ \\
		5 & $2.10$ & $1.84$ & $15.23$ \\
		6 & $2.90$ & $3.00$ & $4.57$ \\
		7 & $1.87$ & $1.17$ & $3.71$ \\
		8 & $1.37$ & $0.98$ & $4.45$ \\
		9 & $1.52$ & $0.91$ & $3.52$ \\
		10 & $1.13$ & $0.83$ & $4.02$ \\
		\midrule
		Mean & $1.69$ & $1.38$ & $5.87$ \\
		\bottomrule
	\end{tabular}
\end{table}

\section{Conclusion}
In this study, an end-to-end methodology for the precise TCP positioning of non-rigid HDLR manipulators using advanced machine vision was presented. Notably, only a relatively low-cost camera was added to the sensor configuration. Emerging technology--related research with heavy-duty machines is still mostly on a proof-of-concept phase, relating to TRL 3, which is partly explained by the multidisciplinary collaborative challenges. Our laboratory-validated results, relating to TRL 4, hold practical relevance and push the boundary toward increased task flexibility and automation level of HDLR manipulators. The proposed methodology for precise TCP positioning is based on computing the OOI position error directly from an image, while utilizing highly accurate motion-based calibration to determine the camera-to-robot relation for a given TCP orientation. The calibration procedure comprised coarse alignment in a global manner by instantiating the VO/SLAM orientations with the initial encoder-based TCP orientation and shifting the translation based on the mass centers of the two point sets, along with final iterative matching.

Many methods in deep neural network--based 6D pose estimation of objects focus on additional pose refinement, which appears to fine-tune the results related to the BOP benchmark datasets. As mentioned earlier, most related studies aim to maximize the performance on these datasets, which is logical because of the competitive nature of the benchmark system. However, our research demonstrated that the bare bones ZebraPose, trained only on synthetic RGB images with no additional pose refinement, has performed excellently in the industry-related application. Despite this, various phenomena, such as occlusion, lighting, reflection, and texture, can present challenges with camera-based measurements depending on the environment. It has been acknowledged that the pose estimation accuracy of instance-level methods has started to saturate, with the focus moving toward methods capable of handling unseen objects, such as pre-trained foundation models. From a practical perspective, visual 6D pose estimation of OOIs would benefit significantly from decreased computation times to potentially enable closed-loop vision-based control. A possible alternative to increase the refresh rate is to incorporate a separate pose tracking network alongside a pose estimation network. However, tracking confidence can decrease over time because of drift, occlusion, or rapid motion, especially in the absence of corrective measures, such as re-initializing using the pose estimation network. Therefore, confidence in pose tracking is perceived as more context dependent than pose estimation.

Resulting from the non-real-time refresh rate of visual 6D pose estimation of the OOI in this work, the vision-based control scheme was realized as point-to-point guidance with several consecutive vision-based reference commands. These commands were not temporally optimized in the presented experiments, but the step-wise execution can be automated for future practical use. The results of each step, namely, the orientation alignment, motion-based calibration, and position alignment, demonstrated sufficient accuracy for the considered application with HDLR manipulators. The presented methodology aims to solve the problem of precise TCP positioning in OOI-focused applications for HDLR manipulators with significant bending, which is a required step toward increased automation and eventual fully autonomous systems. Future research should focus on increasing the TRL in related applications and improving the refresh rate and robustness of visual 6D pose estimation and tracking of OOIs.

\backmatter

\newpage
\bibliography{draft}

\end{document}